\documentclass{article}

\PassOptionsToPackage{numbers, compress}{natbib}
\usepackage[preprint]{neurips_2024}

\usepackage[utf8]{inputenc} 
\usepackage[T1]{fontenc}    
\usepackage{hyperref}       
\usepackage{url}            
\usepackage{booktabs}       
\usepackage{amsfonts}       
\usepackage{nicefrac}       
\usepackage{microtype}      
\usepackage{xcolor}         
\usepackage{graphicx}
\usepackage{subcaption}
\usepackage{amsmath}
\usepackage{bbding}
\usepackage{multirow}
\usepackage{caption}

\title{Zipper: A Multi-Tower Decoder Architecture for Fusing Modalities}

\author{%
  Vicky Zayats\thanks{equal contribution} \quad Peter Chen$^*$ \quad Melissa Ferrari \quad Dirk Padfield\\
  Google DeepMind\\
  \texttt{\{vzayats, chenfeif, melissaferrari, padfield\}@google.com}
}

\begin{document}

\maketitle

\begin{abstract}
Integrating multiple generative foundation models, especially those trained on different modalities, into something greater than the sum of its parts poses significant challenges. Two key hurdles are the availability of aligned data (concepts that contain similar meaning but is expressed differently in different modalities), and effectively leveraging unimodal representations in cross-domain generative tasks, without compromising their original unimodal capabilities.

We propose Zipper, a multi-tower decoder architecture that addresses these concerns by using cross-attention to flexibly compose multimodal generative models from independently pre-trained unimodal decoders. In our experiments fusing speech and text modalities, we show the proposed architecture performs very competitively in scenarios with limited aligned text-speech data. We also showcase the flexibility of our model to selectively maintain unimodal (e.g., text-to-text generation) generation performance by freezing the corresponding modal tower (e.g. text). In cross-modal tasks such as automatic speech recognition (ASR) where the output modality is text, we show that freezing the text backbone results in negligible performance degradation. In cross-modal tasks such as text-to-speech generation (TTS) where the output modality is speech, we show that using a pre-trained speech backbone results in superior performance to the baseline.

\end{abstract}

\section{Introduction}

From text~\cite{brown2020language, anil2023palm, rae2021scaling}, to proteins~\cite{rao2020transformer}, to audio~\cite{borsos2023audiolm}, to images~\cite{yu2022scaling}, to even state sequences~\cite{chen2021decision}, decoder-only generative models have shown that they can be trained to produce useful representations using next-token prediction to successfully generate new sequences in many modalities (e.g. audio, images, or state-action sequences). 
As our world is inherently multimodal, recent works have attempted to create multimodal models capable of generating output in many modalities at the same time~\cite{achiam2023gpt, team2023gemini}. This is usually achieved through some form of vocabulary expansion (converting multimodal representations into discrete tokens and adding them to the base vocabulary of a model) during pre-training or during cross-modal alignment at a later finetuning stage~\cite{yu2024language, kondratyuk2023videopoet}.

While pre-training multimodally comes with strong performance benefits (e.g., a model that natively understands many modalities), it has its drawbacks. For example, it does not solve the problem of how to add a new modality post pre-training. Further, it is inflexible, as adding another modality requires the expensive process of training a new model from scratch and performing hyperparameter search for the optimum training data mixture ratios between the modalities. This makes such a solution ill suited for niche modalities (e.g., IMU, protein sequences, etc.).

Alternatively, vocabulary expansion into another modality can be performed post pre-training on a model that has never seen that modality. As decoder-only models trained in the text modality (e.g., LLMs) have shown extraordinary ability to follow instructions~\cite{ouyang2022training} and learn from examples~\cite{brown2020language} in context, this usually takes the form of grafting another modality, such as audio or image capabilities onto an existing strong text backbone~\cite{rubenstein2023audiopalm, hassid2023textually} via fine-tuning, to take advantage of text modality's expressibility and controllability for human users. However, this also has drawbacks. Namely, the strong text-to-text capabilities of the backbone is destroyed and the resulting model can only perform the cross-modal task it was finetuned for.

Finally in both pre-training and finetuning, large quantities of aligned cross-modal data are required, making both methods ill-suited for modalities where we do not have sufficient quantities of aligned multimodal data.

We propose Zipper, a novel architecture that, at its core, is designed for modularity. Zipper composes multiple unimodally pre-trained decoder-only models. Taking advantage of abundant unsupervised unimodal data, we can pre-train strong decoder-only models in a single modality. We then use cross attention to "zip" together multiple such pre-trained decoders and finetune with limited cross-modal data to enable generative capabilities in multiple modalities. The pre-trained decoder-only models can be flexibly re-used and re-purposed in new multimodal combinations. 

While Zipper architecture can be generalized across multiple modalities and more than two modality backbones, in this work, as a proof-of-concept, we focus on an experimental setup where we fuse only two backbones: speech and text. Empirically, we show strong capabilities of Zipper in simultaneous cross-modal generation for text - Automatic Speech Recognition (ASR) task and speech - Text to Speech Task (TTS). Our experiments on using only a fraction of text-speech aligned data (as low as 1\% of the original data) suggest that first uni-modally pretraining the backbone on unlabeled data  allows Zipper to rely on much less aligned data compared to fine-tuning with vocabulary expansion approaches, providing the possibility that fusing modalities using decoder-decoder architecture may be very useful for generation tasks  with limited amount of paired data. 
To the best of our knowledge, this is the first work that looks into flexible modality composition to enable multimodal generative capabilities by combining separately pre-trained unimodal decoders. 

\hfill \break
\noindent
\textbf{Our contributions are as follows:}

\begin{itemize}
    \item We introduce a new generative multimodal fusion architecture that zips together pre-trained unimodal backbones. Like vocabulary expansion techniques, our proposed architecture can perform generative tasks across all modalities. Unlike vocabulary expansion techniques, our proposed architecture is more flexible and composable, allowing unimodal backbones to be pretrained independently from multimodal alignment finetuning while preserving any unimodality performance (e.g., ensuring no degradation in the text-to-text generation) by freezing the corresponding backbone. 
    \item Based on empirical results on speech and text modalities, we show that our architecture performs competitively for the frozen modality backbone (e.g., text) against vocabulary expansion baseline on text-based generative tasks such as ASR. Further, we show improved word error rate (WER) reduction of 12 absolute points (40\% relative error reduction) on unfrozen modality backbone (e.g., speech) on speech-generative TTS tasks against vocabulary expansion baseline, which we attribute to better alignment capabilities of our cross-attention on long-context generation and finetuning from a strongly pre-trained unimodal speech backbone as a base.
    \item Our ablations suggest that composing unimodally pre-trained models using Zipper is able to learn meaningful representation in scenarios where only a small amount of aligned data is available (as little as 1\%, or under 3K utterances) due to the strong unimodal pre-training of each backbone.
\end{itemize}


\section{Related Work}

Many methods have been explored to bridge multimodal understanding and generation. They can be generally broken down into the broad categories of: vocabulary expansion and encoder-decoder composition. While these methods have shown great promise in their respective downstream tasks (e.g., image-text captioning, ASR, translation, image and video generation, etc.), they require abundant aligned data between modalities for fine-tuning or pre-training. For example, Whisper~\cite{radford2023robust} required 680,000 hours of aligned speech-text data while VideoPoet~\cite{kondratyuk2023videopoet} required 1 billion image-text pairs and 100 million video-text pairs. Such quantities of aligned data may not exist in such quantities or variety for nichier modalities. In our work, we explore whether a unimodally pre-trained decoder backbone (e.g., unimodal data exist in large quantities even in nichier modalities) can potentially ameliorate the need for large quantities of paired data.

\textbf{Vocabulary expansion} techniques generally involve first training useful representations using unsupervised methods and discretizing the embedding space to obtain modality-specific tokens~\cite{chung2021w2vbert, yu2024language}. These tokens are then used to expand the vocabulary of the base LLM either during pre-training or at a later finetuning stage. In vocabulary expansion, data mixtures~\cite{edwards2022translation} are carefully curated and hyperparameters~\cite{aghajanyan2023scaling} carefully selected to ensure modal synergy instead of modal competition. Factors that may affect outcomes include the relative weights of cross-modal task and unimodal task losses, the amount of tokens available for training in each modality and for each modality pairing, training data mixture ratios and model parameter count. While vocabulary expansion is currently the most effective means of achieving multimodal generation, its lack of composability limits its application to nichier modal domains outside of core text, speech and image modalities and as a basis for building models that extend to an arbitrary number of modalities. Further the extensive data curation work and large computation budget requirements (e.g., extensive hyperparameter search and large model sizes needed before synergestic outcomes are observed) is a barrier to entry for all but the largest research organizations. Recent works have also looked at scaling multimodality using mixture-of-experts (MoEs)~\cite{shen2023scaling, lin2024moe}. While they expand the number of parameters in a multimodal model while keeping the number of parameters used for computation constant, they nevertheless face the same limitations. We view such works as orthogonal and tackle the problem of computational efficiency rather than composability.

\textbf{Encoder-decoder composition} has been a core architecture for multimodal learning. One of the most famous examples is Flamingo~\cite{alayrac2022flamingo}, which combined two frozen backbones: text and image using cross-attention and showed emergent capabilities for text generation. Other similar works include~\cite{chen2023pali, yu2022coca, li2023blip} for image-text, ~\cite{tang2023unifying} for document layout-text, ~\cite{radford2023robust} for audio-text and ~\cite{piergiovanni2023mirasol3b} for video-text. Using cross-attention is just one technique for multimodal fusion, an alternative method is to use the adaptor approach where modality representations are directly injected via a simple projection layer~\cite{ma2024embarrassingly} or more specialized connectors~\cite{yu2024connecting, wang2023slm, wu2023decoder} into the decoder-only backbone. Like Flamingo, the text-backbone can also be similarly frozen by using LoRA adapters~\cite{gong2023listen}. While great success in multimodal understanding has been achieved using this kind of architecture, one of its drawbacks is that it can only generate in the backbone modality (e.g., text), whereas our work seeks to enable generative capabilities in any of the unimodally pre-trained backbones. Another interesting aspect to note is that while this line of work focuses on a pre-trained encoder trained in an unsupervised manner, it is unclear whether representations learned by next-token prediction (e.g., unimodally pre-trained decoder) would be superior in some aspects. While we do not explore this in our work, it is an interesting direction to pursue for future work. 

At its core, Zipper fuses two decoder-only backbones in a decoder-decoder compositional setup. While the effect of this on multimodality have not been explored to our knowledge, the more general decoder-decoder architecture have been explored in dGSLM~\cite{nguyen2023generative} and CALM~\cite{bansal2024llm}. In dGSLM, two decoders were fused together for simultaneous generation of dialogue. In CALM, two text-modality decoders trained on different tasks were fused together to introduce new capabilities into the base LLM. The concept is similar to Zipper, but unlike CALM, we explore modality composition rather than task composition within a single modality. 
\section{Model}
\label{sec:arch}

\begin{figure*}
    \centering
    \includegraphics[width=0.9\textwidth]{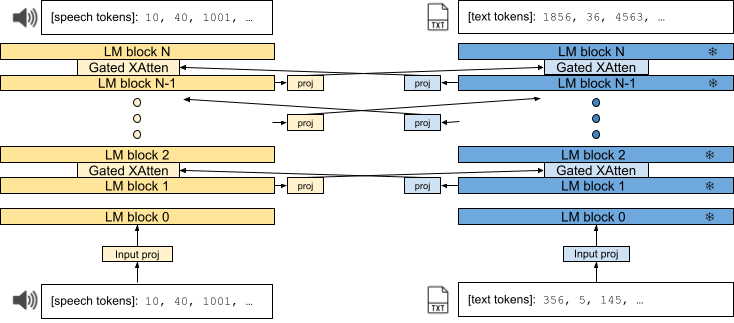}
    \caption{Zipper model with gated cross-attention and projection layers. 
    }
    \label{fig:architecture}
\vspace{-8pt}
\end{figure*}

The Zipper architecture consists of two autoregressive decoder towers (or backbones) that are ``zipped'' together using gated cross-attention layers~\cite{alayrac2022flamingo}. Each backbone is separately trained on a single modality using next-token prediction. Figure \ref{fig:architecture} shows an overview of the Zipper architecture. Similar to CALM~\cite{bansal2024llm}, cross-attention is inserted at every $i$-th layer between the decoder backbones. The representations of a modality at these regularly interleaved layers are cross-attended into the other modality. This differs from Flamingo~\cite{alayrac2022flamingo} encoder-decoder setup, where only the final layer of one tower (an encoder) is cross-attended into the layers of the other (decoder) at regular intervals.

Projection layers are inserted between the modalities during cross-attention. Functionally, this equalizes the embedding dimension size differences between the backbones. Semantically, it also enables the transformation of representations from one modality to another, especially when one or both backbones are frozen. 
Additionally, a non-linear input projection layer is incorporated directly after the input embeddings of each backbone to enable a better adjustment of the unimodal representations of inputs for multimodal tasks.

Let the two autoregressive Transformer decoders be $A$ and $B$. Before the first transformer block (after the embeddings layer), we insert two learnable multilayer perceptron (MLP) projections followed by ReLU transformations for each backbone: 
\vspace{-10pt}

\begin{align*}
    h_B^0 &= ReLU(\text{MLP}_B(\text{embedding}_B(\text{input}_B))) \\
    h_A^0 &= ReLU(\text{MLP}_A(\text{embedding}_A(\text{input}_A))) 
\end{align*}

This is done in order to better adapt the unimodal representations to the multimodal setup. 

Let $i_A$ and $i_B$ represent the intervals at which layers in $A$ are cross-attended into $B$, and $B$ are cross-attended into $A$, respectively. 
We refer to the hidden representation of the unimodal decoder $A$ at layer $k$ as $m_A^k\in\mathbb{R}^{d_A}$ where $d_A$ is the hidden dimension of the transformer $A$ and similarly, to the hidden representation of the unimodal decoder $B$ at layer $l$ as $m_B^l\in\mathbb{R}^{d_B}$ where $d_B$ is the corresponding hidden dimension of transformer $B$. 
Let $f_{\text{cross}}(Q, K, V)$ be a gated cross-attention layer followed by a feed-forward layer from~\cite{alayrac2022flamingo} with $Q, K, V$ being the query, key, and value, respectively.  And let $g^k_A(x)\in\mathbb{R}^{d_A\times{d_B}}$ and $g^l_B(x)\in\mathbb{R}^{d_B\times{d_A}}$ represent the linear feed forward and fully connected projections for towers $A$ and $B$, respectively. 

The new representations $\Tilde{m}_A^k$ at layer $k$ in decoder $A$ are specified as follows:
\begin{align*}
    h_B^l &= ReLU({g^l_B}(m_B^l)) \\
    \Tilde{m}_A^k &= f_{\text{cross}}(Q=m_A^k,K=h_B^l,V=h_B^l) \\ \text{for}&\quad k={n}\cdot{i_A},\quad l={n}\cdot{i_B}, \quad n\in{1,2,3,...}  
\end{align*}
\noindent
Similarly, the new representations $\Tilde{m}_B^l$ at layer $l$ in decoder $B$ are:
\begin{align*}
    h_A^k &= ReLU({g^k_A}(m_A^k)) \\
    \Tilde{m}_B^l &= f_{\text{cross}}(Q=m_B^l,K=h_A^k,V=h_A^k) \\ \text{for}&\quad k={n}\cdot{i_A},\quad l={n}\cdot{i_B}, \quad n\in{1,2,3,...}  
\end{align*}

Finally, each tower ends with  a softmax layer (shared with the same-tower embedding layer) in order to project hidden representations into a probability distribution over the (modality/tower specific) token vocabulary using the next token prediction task.




\subsection{Auto-Regressive Masking}
We adapt the cross-attention mechanism for auto-regressive training on interleaved sequences. This is achieved by cross-attending only to data in the other modality that precedes the current position in the original linear sequence.

\subsection{Inference}
During decoding, the sequence of output modalities is specified (e.g., [speech], [text], [text, speech]). The model generates output in the first modality of the sequence until a special \textit{end-of-sentence} token is encountered, at which point the generation switches to the next modality in the sequence. This process continues until all modalities in the sequence have been decoded. 

While it is possible to extend the model to automatically choose which modality the outputs generate, we leave the generalization of this setup for future work. 
\section{Experiments}
While Zipper can be extended to any number of modalities, all experiments reported here focus on merging text and speech modality. Specifically, we evaluate automatic speech-recognition (ASR) for speech-to-text generation and text-to-speech (TTS) generation.

\subsection{Experimental Setup}
\label{sec:experiment}

In all experiments, variants of PaLM2 \cite{anil2023palm} in two sizes are used as the text backbone. For the text backbone, we use the smallest publicly available PaLM2 Gecko~\cite{PaLM2Gecko} (we refer to as PaLM2-G) model and an even smaller PaLM2 model (we refer to as PaLM2-Tiny).

The speech backbone is based on a similar decoder-only architecture to the one used in PaLM2, with a modified vocabulary size of 1026 (1024 speech tokens and 2 special tokens for beginning and end of audio). The speech backbone is randomly initialized and pre-trained from scratch using the LibriLight\cite{librilight} dataset.
The speech backbone model is available in two sizes: 128M and 1B parameters.  We experiment with three configurations: PaLM2-G/1B speech backbone, PaLM2-Tiny/128M speech backbone and PaLM2-G/128M speech backbone to investigate the effect of model size on our technique. 

Unless mentioned otherwise, we use a total of 4 "zipped" cross-attention layers for PaLM2-Tiny/128M model, 8 "zipped" cross-attention layers for PaLM2-G/128M model, and 9 cross-attention layer for PaLM2-G/1B model. For all the projection layers we use a 3-layer MLP. We ablate aspects of our model design in Appendix~\ref{sec:app}.

We trained our models using 16 TPU v3 chip for 1M steps with batch size of 512. In all experiments we leave the text backbone frozen and experiment with a frozen and unfrozen speech backbone. The reason is that in most cases, the text backbone is extremely strong compared to all other modalities and in most applications we would like to preserve the text-to-text capabilities while infusing generative capabilities in another modality, such as speech. Unlike encoder-decoder architectures like~\cite{wang2023slm, wu2023decoder, yu2024connecting, ma2024embarrassingly}, zipper has the ability to perform cross-modal tasks such as TTS, alongside uni-modal and cross-modal text generation capability.

\subsubsection{Data}
\label{sec:eval_data}
Besides pre-training of the speech backbone, which is done on unlabeled LibriLight dataset\cite{librilight} that consist of 60,000 hours of data, all our models are trained on a mixture of ASR (speech-to-text) and TTS (text-to-speech) tasks using a combination of the LibriSpeech \cite{panayotov2015librispeech} and LibriTTS~\cite{zen2019libritts} datasets with ``clean'' (100 hours, clean), ``other-360'' (360 hours, noisy), and ``other-500'' (500 hours, noisy) data splits. The datasets are mixed based on the number of total 
samples in each dataset, which results in close to 1:1.26 mixing ratio between ASR and TTS tasks.

\subsubsection{Baseline}
For baseline, we use a single-tower decoder (which we refer to as Single Decoder) consisting of a pre-trained PaLM2 backbone that had its vocabulary extended with an extra 1026 semantic speech tokens. We fine-tuned on the same ASR and TTS tasks as used in Zipper experiments. The setup is similar to AudioPaLM\cite{rubenstein2023audiopalm}, but fine-tuning only with publicly available LibriSpeech and LibriTTS datasets to enable a fair comparison with Zipper. We experiment with the same two model sizes as the text-side backbone of Zipper: PaLM2-G and PaLM2-Tiny. While this baseline is far from state-of-the-art, it provides an minimally-viable apple-to-apples comparison with publicly accessible datasets.
\vspace{-5pt}

\subsubsection{Speech Tokenization and Generation}
\label{sec:token_gen}
We follow the same procedure as SoundStorm~\cite{borsos2023soundstorm} to obtain speech (semantic) tokens using quantized w2v-BERT\cite{chung2021w2vbert} embeddings. The vocabulary size of these semantic speech tokens is 1024. The w2v-BERT encoder is off-the-shelf and part of the speech pre-processing pipeline. It was pre-trained on LibriLight\cite{librilight} (60,000 hours) dataset using unsupervised constrastive learning and masking objectives. These speech tokens are then used to train Zipper and our decoder-only baseline in a next-token prediction setup. 

For speech generation (TTS), Zipper and the decoder-only baseline will generate the semantic speech tokens. These tokens are then passed through the SoundStorm decoder to convert the semantic tokens back into audio. 
\vspace{-5pt}

\subsubsection{Evaluation}
\label{sec:eval}
We evaluate cross-modality generation for text on ASR tasks where we use Word Error Rate (WER) as the metric. 

For cross-modality generation for speech, we evaluate on TTS tasks where the generated speech is then passed through an unrelated ASR system (\textsc{Eval ASR System}~\cite{zhang2023google}) and also evaluated using WER as the metric. 

While the more standard evaluation of TTS systems (synthesized speech) rely on human feedback (Mean Opinion Score) that capture many holistic aspects of speech (e.g., fidelity to text and acoustic quality, etc.), in our TTS evaluation, we wish to solely capture the impact of the choice of architecture have on the ability to model and predict semantic tokens. Therefore, as the primary task of semantic tokens is to capture content information, we proxy the quality of the generated semantic tokens with its fidelity and alignment to the desired gold-transcript. Other aspects of speech (e.g., acoustic quality), we consider out-of-scope in our evaluation and we find to be more artifacts of the quality of the decoder (SoundStorm\cite{borsos2023soundstorm}) which in all setups, we hold the same across our experiments. 

Specifically, the TTS evaluation pipeline is as follows. Zipper and associated baseline models generate semantic tokens based on text conditioning. These semantic tokens are passed to SoundStorm decoder as \textit{conditional} tokens (as described in original paper) and SoundStream~\cite{zeghidour2021soundstream} acoustic tokens are generated. These acoustic tokens are then decoded into audio using the off-the-shelf SoundStream decoder. The audio is passed to our \textsc{Eval ASR System} for transcript prediction and the WER is calculated. 

We calculate the WER with respect to the gold transcript and to the target transcript from our \textsc{Eval ASR System} on audio generated by SoundStorm from gold speech tokens. In this way we can differentiate between absolute end-to-end performance and relative performance against a baseline with oracle semantic token prediction.

Finally, one potential pathology of our automatic metric is that the decoder is so weak (that it generates only noisy speech) or \textsc{Eval ASR System} is so error prone, that it prevents us from capturing the true differences between our baseline and our proposed method. To sanity check that our choice of decoder and \textsc{Eval ASR System} is suitable, we provide the absolute oracle WER of gold speech tokens with respect to gold transcripts as a measure of performance ceiling due to error propagation caused by by SoundStorm speech decoder and \textsc{Eval ASR System}. We find that the performance ceiling is still significantly above our results from Zipper and baseline.
\vspace{-5pt}

\subsubsection{Significance scores}
\label{sec:sig}
In all our main experiments on ASR and TTS tasks (Sections~\ref{sec:asr},\ref{sec:tts}) we report significance scores with respect to the corresponding vocabulary expansion baseline (e.g., same sized text backbone) using Wilcoxon signed-rank test calculated on a sentence level data points. We indicate the \textit{p-value} used for significance calculation in the corresponding table caption ($p<0.01$ or $p<0.05$).

\subsubsection{Hyperparameter Tuning}
\label{sec:hyper}

 The Single Decoder model used Adafactor optimizer with a constant learning rate, dropout rate of 0.1. 
 Zipper used gradient clipping with a max norm of 1.0, and Adam optimizer with $\beta_{1}=0.9$ and $\beta_{2}=0.99$. The speech backbone was pre-trained for 400M steps with the batch size of 1024 and learning rate of 5e-4.

We performed a hyperparameter search over the learning rates 5e-5, 1e-4, 5e-4, 1e-3 for finetuning both Zipper and Single Decoder. The Single Decoder has the best performance with the learning rate of 1e-4, while Zipper has optimal performance at a higher learning rate of 5e-4 when the speech backbone is unfrozen and 1e-3 when the speech backbone is frozen. Hyperparameter search is done on ASR task using the validation split based on the geometric mean of WER calculated on \textit{clean} and \textit{other} subsets. Unless mentioned otherwise, all final results in Sections~\ref{sec:asr},~\ref{sec:tts} are reported on test splits: \textit{clean} and \textit{other} on ASR task, and \textit{clean} on TTS task.


\subsection{Automatic Speech Recognition}
\label{sec:asr}


\begin{table}[!t]
\centering
\footnotesize
\setlength{\tabcolsep}{9pt}
\begin{tabular}[t]{|l|c|c|c|c|} \hline
\textbf{Model} & Size (text/speech) & Frozen(\SnowflakeChevron) & test-clean & test-other \\ \hline
SLAM-ASR w/ Whisper Enc~\cite{ma2024embarrassingly} & 1B/1.5B& yes & 4.33 & 8.62 \\ \hline
\multirow{2}{*}{SLAM-ASR w/ HuBERT Enc~\cite{ma2024embarrassingly}} & 7B/316M& yes & 2.30 & 4.53 \\ \cline{2-5}
 & 7B/1B & yes & 1.94 & 3.81 \\ \hline
 Q-Former connector w/ Whisper~\cite{yu2024connecting} & 7B/1.5B& yes & 2.30 & 4.53 \\ \hline \hline

\multirow{2}{*}{Single Decoder} & Tiny & - & 3.78 & 7.53 \\ \cline{2-5}
 & Gecko & - & 3.49 & \textbf{7.09} \\ \hline \hline
\multirow{6}{*}{Zipper} & \multirow{2}{*}{Tiny/128M} & no & 3.54 & 8.34**\\ \cline{3-5}
 &  & yes & 3.47* & 8.37** \\ \cline{2-5}
 & \multirow{2}{*}{Gecko/128M} & no & 3.31* & 7.70**\\ \cline{3-5}
 &  & yes & 3.36* & 7.90** \\ \cline{2-5}
 & \multirow{2}{*}{Gecko/1B} & no & 3.17 & 7.42** \\ \cline{3-5}
 &  & yes & \textbf{2.95} & 7.15 \\ \hline
\end{tabular}

\caption{WER of the Single Decoder vs Zipper on LibriSpeech ASR task on \textit{test-clean} and \textit{test-other} splits. 
Single asterisk indicate when Zipper model is significantly better than Single Decoder baseline, double asterisk indicate when Zipper model is significantly worse than Single Decoder baseline, where the significance p-value is $p<0.05$ in both cases.}
\label{tab:asr}
\vspace{-24pt}
\end{table}

The results for the ASR task on the test splits are presented in Table~\ref{tab:asr}. While this work focuses on models that are able to generate across all the modalities of its unimodal pre-trained backbones, we also include recent encoder-decoder results from~\cite{yu2024connecting} and~\cite{ma2024embarrassingly} for comparison, however these models can only output in the decoder modality (text). In~\cite{ma2024embarrassingly}, Whisper~\cite{radford2023robust} (pre-trained) and HuBERT~\cite{hsu2021hubert} (pre-trained and finetuned) encoders were fused with a Vicuna~\cite{chiang2023vicuna} LLM via linear projectors. Similarly, in~\cite{yu2024connecting}, the authors fused an Whisper encoder finetuned on ASR tasks using Q-Former connectors. 

When comparing the Zipper to the vocabulary expanded Single Decoder baseline, we observe that Zipper has slightly better performance on \textit{test-clean} subset, and comparable to slightly-degraded performance on the noisier speech \textit{test-other} subset. We observed that freezing the speech backbone yields slightly better performance in most of the settings on \textit{test-clean} and slightly worse performance on \textit{test-other}, although overall the performance is still quite similar. 
It is worth noting that while in the Single Decoder setup all of the parameters are updated, in Zipper we either freeze both backbones and only allow cross-attention weights to update or we allow both cross-attention weights and speech backbone to update. This setup by definition ensures that text-to-text capabilities of the text backbone do not degrade. These results show that this can be achieved at almost no cost to cross-modal generative tasks (ASR) in the frozen domain itself. 

Also, consistent with findings across a variety of modalities, including speech and text, we found that larger model sizes lead to better performance. This findings were similarly found to be true in both~\cite{ma2024embarrassingly} and~\cite{yu2024connecting}. This in part explain the gap in performance between the larger 7B text backbone results for the encoder/decoder model against the much smaller models we used for our Single Decoder and Zipper experiments. 
 




\subsection{Text to Speech}
\label{sec:tts}


\begin{table}
\centering
\footnotesize
\setlength{\tabcolsep}{9pt}
\begin{tabular}[t]{|l|c|c|c|} \hline
\textbf{Model} & Frozen(\SnowflakeChevron) & WER (w.r.t. gold) & WER (w.r.t. target)  \\ \hline
Oracle (target) & - & 6.00 & - \\ \hline \hline
Single Decoder (T) & - & 33.38 & 33.11 \\ \hline 
Single Decoder (G) & - & 32.35 & 31.99 \\ \hline \hline 
Zipper T/128M & no & 20.14* & 19.45* \\ \hline 
Zipper T/128M & yes & 27.87 & 27.33* \\ \hline 
Zipper G/128M & no & 17.70* & 16.96* \\ \hline 
Zipper G/128M & yes & 22.48* & 21.81* \\ \hline 
Zipper G/1B & no & 19.97* & 19.21* \\ \hline 
Zipper G/1B & yes & 20.35* & 19.71* \\ \hline 
\end{tabular}
\caption{Performance of Single Decoder vs Zipper on \textit{test-clean} split of LibriTTS. WER (gold) refers to WER with gold transcript as reference, and WER (target) refers to WER w.r.t. the target transcript generated from gold speech tokens. Oracle corresponds to the WER of the gold speech tokens with respect to gold transcript. 
Asterisk indicate significance at $p<0.01$ using Wilcoxon signed-rank test. 
}
\label{tab:tts}
\vspace{-12pt}
\end{table}


The results for the TTS task on the \textit{test-clean} split of the LibriTTS dataset are presented in Table~\ref{tab:tts}. We only report results on the non-noisy \textit{test-clean} split, as our TTS system is unlikely to be able to replicate noise, making comparison based on our \textsc{Eval ASR System} unfair for the noised \textit{test-other} split. We report WER of the generated speech with respect to two references: the gold transcript WER (gold), and the target transcript WER (target), as explained in Section~\ref{sec:eval}. To obtain more consistent metrics, for all results in Table~\ref{tab:tts} we run SoundStorm decoder and \textsc{Eval ASR System} 3 times (with different random seeds), and report the best out of 3 WER for each sample. 

Zipper models significantly outperform Single Decoder models, leading to 13 WER points improvement (40\% relative error reduction) for Zipper S/128M unfrozen models and 12 WER point improvement (38\% relative error reduction) for Zipper L/1B unfrozen models. We also observe that unfreezing the speech backbone during training consistently improves performance across all Zipper model sizes compared to using a frozen backbone, which matches the general intuition that finetuning the parameters of the speech backbone should produce better modal alignment than relying solely on cross-attention.

As discussed in~\ref{sec:eval}, we also report ``Oracle'' WER of the transcript generated from the gold speech tokens with respect to the gold transcript as a performance ceiling and to show that despite error propagation from SoundStorm decoder and \textsc{Eval ASR System}, our automatic metric is sufficiently robust that it is not a bottleneck in evaluation. 

\begin{figure}
\centering
\begin{minipage}{.45\textwidth}
  \centering
  \includegraphics[width=0.97\linewidth]{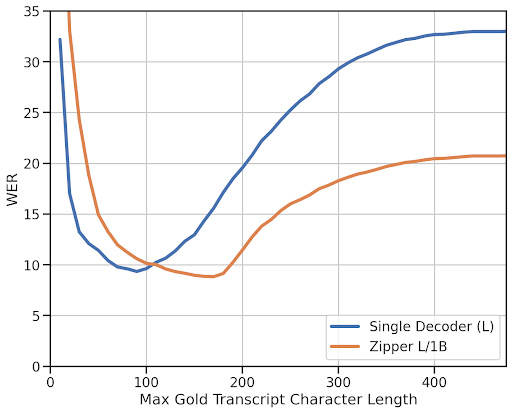}
  \captionof{figure}{WER on TTS task from Zipper and Single Decoder models vs. max gold transcript length.}
  \label{fig:goldtranscript}
\end{minipage} \qquad
\begin{minipage}{.45\textwidth}
  \centering
  \includegraphics[width=1\linewidth]{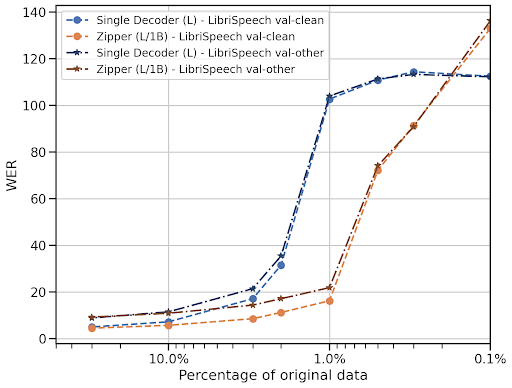}
  \captionof{figure}{WER on ASR task (validation set) as a function of the amount of aligned data.}
  \label{fig:data}
\end{minipage}
\end{figure}
\vspace{-8pt}

We hypothesize that the large gains in TTS may be first due to the opportunity to fully leverage a pre-trained speech backbone, whereas in the Single Decoder setup, a new modality must be integrated into a pre-trained text model with only finetuning cross-modal data. Second, we believe that cross-attention might have an advantage over self-attention especially for longer outputs (speech) due to its ability to provide a more global view of the cross-modal (text) context. 
As shown in Figure~\ref{fig:goldtranscript}, we find Zipper is able to primarily outperform Single Decoder setups in longer-TTS generation scenarios while underperforming for shorter utterances where sequence length of the generated speech is less of a factor. 

\subsection{Amount of Aligned Data}
\label{sec:align_data}




Composing multimodal models from unimodal decoders may be especially difficult in multimodal domains with limited amounts of aligned data. We investigate this aspect by varying the amount of data used for training. Specifically, we show WER performance for ASR task on validation set trained with only a small subset of the original aligned data (mixture of LibriSpeech and LibriTTS). We vary the training data quantity from 0.1\% to 30\% of the original data size. 

The results for the clean and noisy data subsets are shown in Figure~\ref{fig:data} where the x-axis shows the fraction of aligned data on logarithmic scale. The results suggest that Zipper achieves much better results with significantly less aligned data. Notably, when only 1\% of the original data is used (corresponding to only 2.8K training utterances for ASR task) for training/fine-tuning, Zipper is capable of learning meaningful speech-to-text mappings, with a WER in the mid-twenties compared to the Single Decoder setup with a WER around one hundred. 

We believe the improvement demonstrated with Zipper is due to the use of a strong pre-trained speech backbone, enabling the model to leverage the unlabeled speech data on which it was pre-trained to overcome the limitations of the lack of aligned data.

\section{Conclusion and Future Work}


In this paper, we introduced Zipper, a multi-tower decoder architecture for composing independently pre-trained unimodal decoders to enable multimodal generative capabilities. Our method allows each modality to independently retain its unimodal generative capabilities (e.g., keeping their parameters frozen during cross-modal alignment) if desired.

Our experiments on zipping together speech and text modalities
demonstrate competitive cross-modal performance on the frozen modality (e.g., text generation on ASR tasks) and absolute WER reduction of 12 points (relative WER reduction of 40\%) on the unfrozen modality (e.g. speech generation on TTS tasks) compared to the baseline/traditional approach of expanding the vocabulary (e.g., with speech tokens) and cross-modaly finetuning a text model. 
We also show that, by enabling the (re)use of strongly pre-trained unimodal models, Zipper is capable leveraging these as backbones to learn with limited aligned data, indicating the usefulness of the approach in extreme cross-modal data-constrained scenarios. 

For future work, we aim to extend the model beyond two unimodal decoders to demonstrate how it can be used to combine a larger number of modalities (e.g., jointly understanding and generating in modalities such as text, speech, video, images, etc.). We also plan to scale Zipper to larger model sizes and greater diversity of data.


\section{Limitations}
\label{sec:limit}
This paper presents preliminary work on modular fusion of unimodally pre-trained backbones. As the main focus of this paper is a proof-of-concept on the new multimodal architecture, therefore we only focus on fusing the text and speech modalities. In order to evaluate the generality of the proposed architecture across a variety of modalities, further empirical evidence involving other modalities, including more niche modalities, with more diverse set of tasks would be required. Furthermore, the experimental setup used in this paper is limited and small scale. Our model sizes are small, and data is limited only to academic datasets on read speech. Finally, while focusing on the modular nature of the approach, we did not investigate fully the possible architectural components of the model, such as using a shared vs. domain specific MLP layer in cross-attention, or extensively experimenting with other layers or activations. Finally, we only experimented with fusion of two modalities. Although our architecture could be extended to three or more modalities, we left unexplored the topic of whether more than two modalities can be fused using only bimodally aligned data as trimodally aligned data is even more rare.




\begin{ack}
We would like to thank Chulayuth Asawaroengchai and Duc Dung Nguyen for their help with early experimentation, Zalán Borsos, Alex Tudor and Daria El Badawy for their help with AudioLM and SoundStorm pipelines, and Christian Frank, Jesper Andersen, and Rif A. Saurous for their advice and support. Their support greatly contributed to the success of this research.

\end{ack}

\medskip

{
\small
\bibliographystyle{plain}
\bibliography{mybib}
}


\appendix
\newpage
\section{Appendix}
\label{sec:app}

We ablate various aspects of our model architecture: input projection layers and number of cross-attention layers to better the impact of certain model design choices. 

We perform our analysis on the validation split of \textit{clean} data, also referred as \textit{val-clean} on PaLM-Tiny/128M Zipper model with unfrozen speech backbone. We use ASR and TTS tasks as  evaluation benchmark and WER as our metric following the same methodology as defined in Section~\ref{sec:eval} (WER).

\begin{figure}[t]
    \begin{subfigure}{0.5\textwidth}
        \centering
        \includegraphics[width=0.97\linewidth]{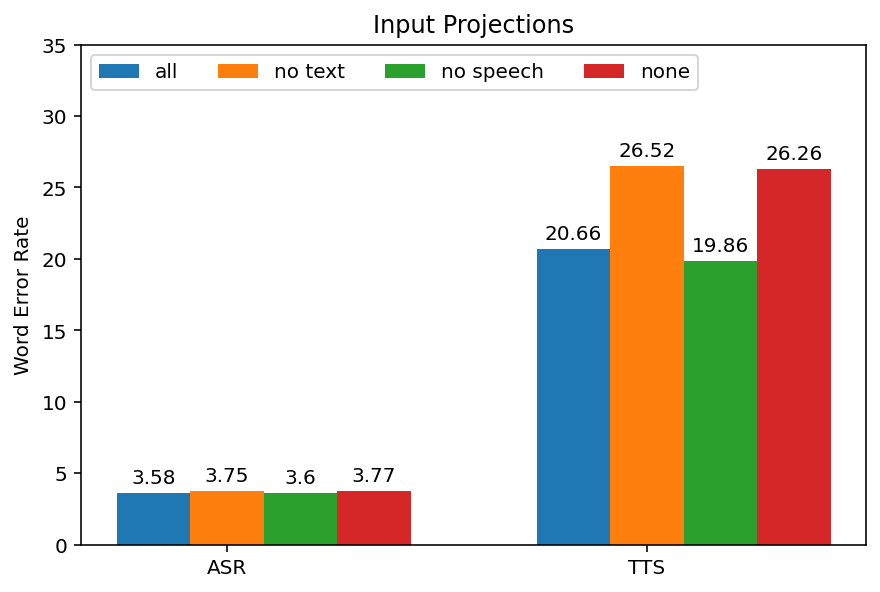}
        \caption{Model ablation on input projection layers.}
        \label{fig:input_proj}
    \end{subfigure}
    \begin{subfigure}{0.5\textwidth}
        \includegraphics[width=1\linewidth]{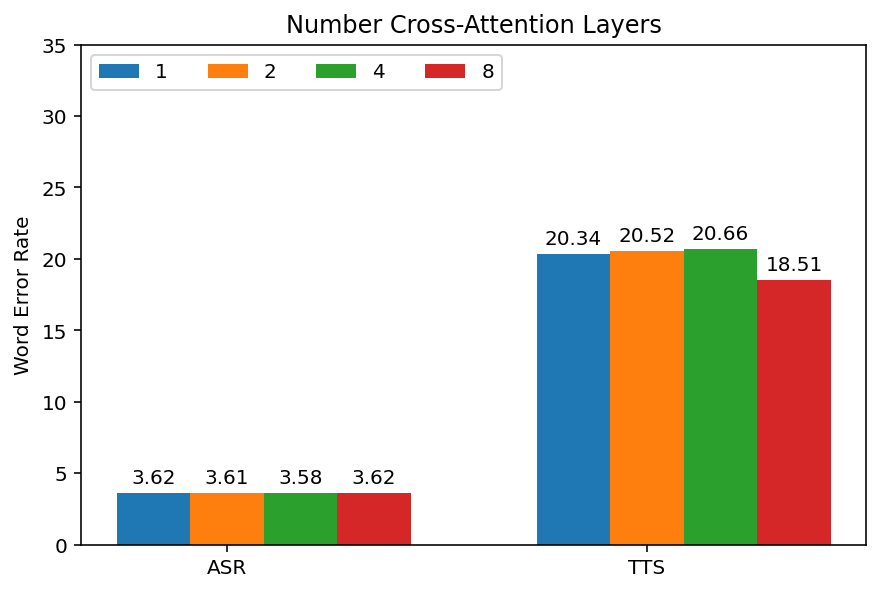}
        \caption{Affect of the number of cross-attention layers.}
        \label{fig:cross}
    \end{subfigure}
    \caption{Model design - ablations with respect to input projections and number of cross-attention layers. }
\end{figure}

Figure~\ref{fig:input_proj} shows the ablation for input projection where either the text input projection, speech input projection, or both input projections are removed. We see that input text projection is especially important, especially for TTS performance. We believe the text input projection is more important because we held the text backbone frozen, whereas the speech input projection is less important because we unfroze the speech backbone itself to be finetuned, potentially obviating the need for projection layers to adapt modal representations. 

Figure~\ref{fig:cross} shows ablation on the number of cross-attention layers used to fuse the two backbones. The maximum number of layers (8) improves TTS performance with not much effect on ASR performance. Counter-intuitively, we saw no significant change in performance when smaller number of cross-attention layers are used.

\end{document}